\documentclass[runningheads]{llncs}

\usepackage{eccv}

\usepackage{eccvabbrv}

\usepackage{graphicx}
\usepackage{booktabs}
\usepackage{multirow}
\usepackage[accsupp]{axessibility}  

\usepackage[pagebackref,breaklinks,colorlinks,citecolor=eccvblue]{hyperref}

\usepackage{orcidlink}

\begin{document}

\title{Rate-Distortion-Cognition Controllable\\ Versatile Neural Image Compression} 


\author{Jinming Liu$^1$$^,$$^2$, Ruoyu Feng$^3$, Yunpeng Qi$^3$, Qiuyu Chen$^2$, Zhibo Chen$^3$, Wenjun Zeng$^2$, Xin Jin$^2$$^\dag$
}

\authorrunning{Jinming Liu et al.}

\institute{Shanghai Jiao Tong University \and
Ningbo Institute of Digital Twin, Eastern Institute of Technology, Ningbo, China \and University of Science and Technology of China\\
\email{jmliu206@sjtu.edu.cn\quad jinxin@eitech.edu.cn}}

\maketitle
\let\thefootnote\relax\footnotetext{\dag~Corresponding author.}

\begin{abstract}
\vspace{-5mm}
  Recently, the field of Image Coding for Machines (ICM) has garnered heightened interest and significant advances thanks to the rapid progress of learning-based techniques for image compression and analysis. Previous studies often require training separate codecs to support various bitrate levels, machine tasks, and networks, thus lacking both flexibility and practicality. To address these challenges, we propose a rate-distortion-cognition controllable versatile image compression, which method allows the users to adjust the bitrate (i.e., Rate), image reconstruction quality (i.e., Distortion), and machine task accuracy (i.e., Cognition) with a single neural model, achieving ultra-controllability. Specifically, we first introduce a cognition-oriented loss in the primary compression branch to train a codec for diverse machine tasks. This branch attains variable bitrate by regulating quantization degree through the latent code channels. To further enhance the quality of the reconstructed images, we employ an auxiliary branch to supplement residual information with a scalable bitstream. Ultimately, two branches use a `$\beta x + (1 - \beta) y$' interpolation strategy to achieve a balanced cognition-distortion trade-off. Extensive experiments demonstrate that our method yields satisfactory ICM performance and flexible Rate-Distortion-Cognition controlling.
\end{abstract}

\vspace{-10mm}
\section{Introduction}
\vspace{-2mm}
\label{sec:intro}
Machine vision tasks have become increasingly integral in numerous daily applications, including intelligent traffic systems~\cite{gao2021digital}, autonomous driving~\cite{hu2023planning}, and facial recognition~\cite{terhorst2023qmagface}.
As a result, there is a burgeoning consumption of images by machines, which poses significant challenges in the transmission and storage of such extensive data volumes. How to efficiently compress these data while promising machine task accuracy urgently needs to be studied.

Traditional image compression standards, such as JPEG~\cite{wallace1992jpeg}, HEVC~\cite{sullivan2012overview}, VVC~\cite{bross2021overview}, have been widely applied in life. Meanwhile, deep learning-based codecs have rapidly developed recently, achieving superior rate-distortion performance~\cite{liu2023learned,cheng2020learned,lu2021transformer, li2023frequency}. However, despite success in bit saving and visual fidelity, these methods hardly generalize well to machine tasks w.r.t cognition performance. 

That's because traditional image compression codecs are typically optimized to align with human visual perception. They employ distortion metrics, such as PSNR and MS-SSIM~\cite{liu2023learned}, as well as realism metrics, including KID and FID ~\cite{agustsson2023multi}, to assess the loss incurred by compression procedure. Hence, utilizing these codecs directly for machine vision tasks may yield suboptimal outcomes~\cite{duan2020video,lu2022preprocessing}.

\begin{figure}[t]
    \centering
    \includegraphics[width=1\linewidth]{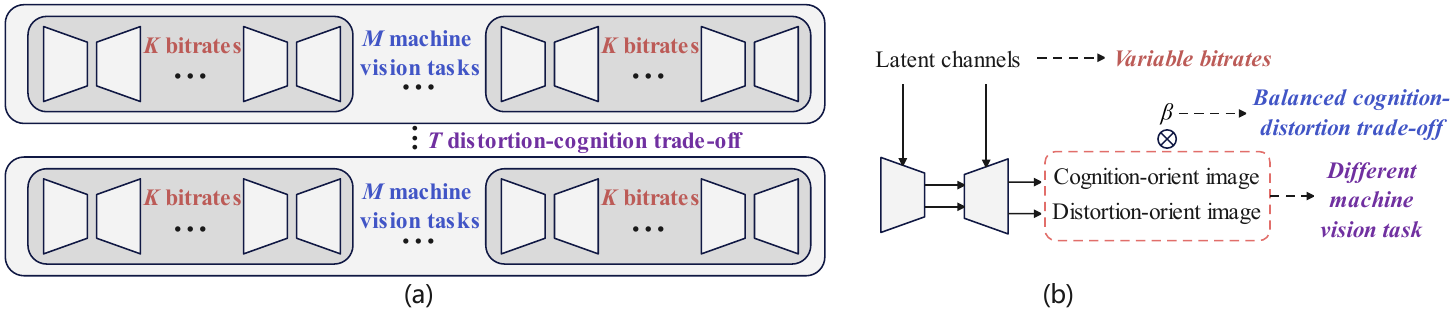}
    \vspace{-6mm}
    \caption{Motivation: (a) Traditional image compression tasks require training $K \times M \times T$ codecs for $K$ bitrate points, $M$ machine vision tasks, and $T$ cognition-distortion trade-offs. (b) Our method enables controllable rate-distortion-cognition with a single codec.}
    \vspace{-5mm}
    \label{fig:head}
\end{figure}

To address this issue, the field of image coding for machines (ICM) has gradually come into focus~\cite{he2019beyond,duan2020video,sun2020semantic,jin2023semantical,li2024image}. For example, some works improves the performance of downstream tasks by jointly optimizing the codec and the task network~\cite{ma2020end}. Choi \etal~\cite{choi2022scalable} adopt a scalable method to decouple the features required for downstream machine vision tasks and reconstruction tasks, thereby reducing the required bitrate. Chen \etal~\cite{chen2023transtic} fine-tune the codec for different tasks through prompts to achieve better performance. Feng \etal~\cite{feng2022image} employ a contrastive learning approach to learn an omni-feature to improve the performance on various downstream tasks. 
However, these methods all remain the following challenges: 1). need to train multiple independent networks to adapt to different bit rates, thus the training cost and memory requirement increase with the desired rate-distortion points proportionally. 2). hard to accommodate diverse machine tasks/networks, bringing extra fine-tuning overhead and bad generalization. 3). fail to meet the practical demand for adjustable distortion-cognition trade-off (decompression fidelity vs. task performance) within a single framework.
In summary, as illustrated in Fig. \ref{fig:head}(a), previous image compression methods require training different codecs to support various rate-distortion levels and machine tasks, posing a large overhead in practical scene deployment. 

Some variable-bitrate (VBR) image compression methods attempt to achieve compression at different bitrates through a single model~\cite{song2021variable,cui2020g}. For example, Song \etal~\cite{song2021variable} use a quality map to indicate compression bitrates, achieving compression at various rates. Cui \etal~\cite{cui2020g} achieve VBR by scaling different channels of the latent code through a gain unit. However, to our best knowledge, there are currently no VBR methods specifically designed for machine tasks. Most VBR methods are optimized for human vision, and can not be directly applied to machine vision tasks due to the assessment gap.

Meanwhile, the \emph{distortion-realism} trade-off of compression has been discussed recently~\cite{agustsson2023multi,korber2023egic}. For example, Agustsson \etal~\cite{agustsson2023multi} achieve adjustable distortion-realism by attaching several layers of neural networks outside the decoder. However, both evaluations of \emph{distortion} and \emph{realism} align with human perception intuition, and many assessment aspects of them are largely overlapped. Unlike \emph{distortion} or \emph{realism}, \emph{cognition} requires more valid information needed for downstream tasks. To the best of our knowledge, there are currently no studies discussing how to adjust the distortion-cognition trade-off in a single codec.

\begin{figure}[t]
    \centering
    \includegraphics[width=1\linewidth]{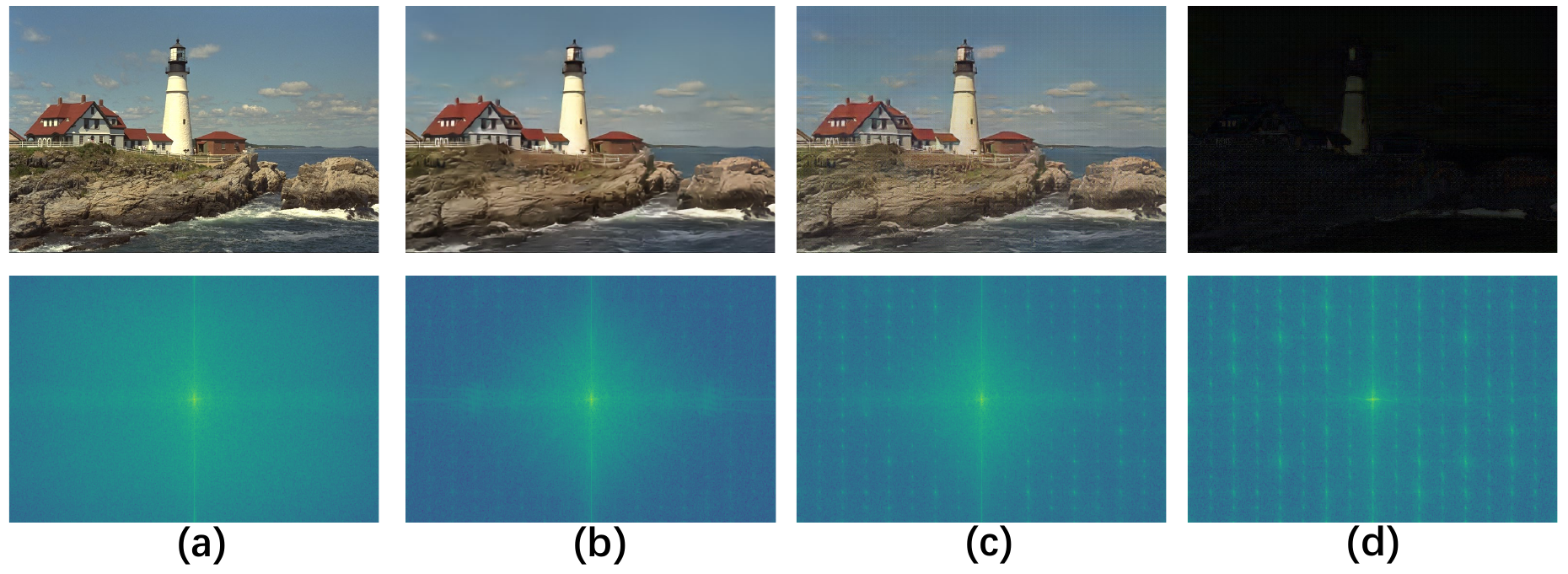}
    \vspace{-7mm}
    \caption{The visualization and spectrum of (a) original image, (b) distortion-oriented compressed image $\boldsymbol{\hat{x}}_2$, (c) cognition-oriented compressed image $\boldsymbol{\hat{x}}_1$ and, (d) the residual $|\boldsymbol{\hat{x}}_1 - \boldsymbol{\hat{x}}_2|$ of (b) and (c). 
    }
    \vspace{-6mm}
    \label{fig:example}
\end{figure}

In response to the aforementioned issues, we propose a rate-distortion-cognition controllable versatile neural image compression method. Specifically, we first introduce a cognition-oriented loss that combines contrastive constraints in the main compression branch to reconstruct a more general machine task-friendly image. The physical meaning behind this loss is that contrastive learning enables better visual representations for downstream intelligent analysis tasks~\cite{he2020momentum,caron2021emerging,chen2020simple}. We can see from Fig. \ref{fig:example} that, the normal distortion-oriented compressed images contain more low-frequency information, aligning with the intuition that the human eye is more sensitive to low-frequency components of image~\cite{yan2017efficient}. In contrast, the cognition-oriented compressed images contain more valid and critical high-frequency information, making them more suitable for machine vision analysis. 

Then, we attain variable rate-cognition by regulating quantization degree based on a channel-controllable latent gain unit~\cite{cui2020g}. This module is end-to-end trainable and can be well compatible with the above contrastive objectives. Moreover, we analyze the difference in the information needed for cognition and distortion, and based on the results we further propose an auxiliary branch to supplement residual information with an extra scalable bitstream. We find only few bits cost could significantly reduce the distortion in compressed images. At last, by controlling interpolation between distortion-oriented and cognition-oriented compressed images, we can adjust the distortion-cognition trade-off in a single codec, thereby achieving controllable rate-distortion-cognition. This effectively enhances the practicality of ICM methods.

The contributions of this paper can be summarized as follows:
\begin{itemize}

    \item We propose a rate-distortion-cognition controllable versatile neural image compression model. 
    To the best of our knowledge, we are the first to achieve controllable rate-distortion-cognition within a single codec.
    
    \item We introduce a cognition-oriented loss based on contrastive learning to generate task-friendly compressed images, which main compression process also attains variable-bitrate with a trainable gain unit.
    
    \item We explore the relationship between the information required for distortion and cognition, and design an auxiliary scalable bitstream to achieve a balanced cognition-distortion trade-off with a two-branch interpolation strategy.
\end{itemize}

Experimental results demonstrate that our approach achieves satisfactory performance in both machine tasks and reconstruction, and outperforms both the latest ICM and learned image compression (LIC) methods in terms of adjustable rate-distortion-cognition practicality.

\vspace{-3mm}
\section{Related Works}
\vspace{-2mm}
\subsection{Image Compression}
\vspace{-1mm}
Image compression seeks to encode original image in a format that is both compact and retains high fidelity. Traditional image codecs, as delineated in numerous studies \cite{wallace1992jpeg,rabbani2002overview,wiegand2003overview,sullivan2012overview,bross2021overview}, typically incorporate modules such as intra prediction, discrete cosine transformation or wavelet transformation, quantization, and entropy coding. In contrast, learned-based codecs \cite{balle2017end,minnen2018joint, cheng2020learned,mentzer2018conditional,mentzer2020high,liu2023learned} leverage neural networks to minimize distortion and bitrate. Compared to traditional approaches, data-driven learned methods have currently achieved better rate-distortion performance.
 Moreover, the rate-distortion trade-off in these models is governed by a Lagrange multiplier $\lambda$, which typically limits most methods to a single operation point on the rate-distortion curve for a fixed $\lambda$ value. Recent advancements\cite{choi2019variable,cui2020g,johnston2018improved,toderici2017full,yang2020variable} have introduced various approaches to enable variable-rate compression within a single model framework. For example, Song \etal \cite{song2021variable} innovatively propose spatial bit allocation in accordance with a quality map that matches the dimensions of the original image, offering a novel direction in the field of image compression. Cui \etal \cite{cui2020g} utilze a scale matrix to scale the different channels of latent code to achieve variable bitrates. But these methods can not be directly applied to machine vision tasks due to the assessment gap.

\vspace{-2mm}
\subsection{Image Compression for Machine}
\vspace{-1mm}
Image Coding for Machines (ICM) focuses on compressing and transmitting source images to facilitate downstream intelligent tasks, including image classification \cite{jia2022visual,han2021transformer,dosovitskiy2020image}, object detection \cite{ren2015faster,redmon2016you, redmon2017yolo9000,li2022exploring}, semantic segmentation \cite{long2015fully,badrinarayanan2017segnet,chen2017deeplab,xie2021segformer,zheng2021rethinking}. Joint optimization of image compression models with these intelligent tasks \cite{akbari2019dsslic,hu2020towards,li2021task,le2021image,wang2021towards} is a straightforward solution in this domain. Some feature compression methods are also proposed. \cite{duan2015overview,duan2018compact,chen2019lossy,chen2019toward,singh2020end,ma2018joint,babenko2014neural,feng2022image}, enhancing both coding efficiency and computing offloading. For example, Feng \etal \cite{feng2022image} try to compress an omni-feature for better machine vision task performance. Similarly, methods based on feature selection have been proposed~\cite{liu2023learned,liu2022semantic, feng2023prompt, chen2023transtic, sun2020semantic}, achieving lower compression rates by compressing local regions instead of all features/images. For instance, Chen~\cite{chen2023transtic} \etal fine-tunes the codec via prompts, making the compressed features more suitable for downstream tasks. However, the methods mentioned above require training a separate model for compression at different bitrate points, and currently, there has been no discussion on variable-bitrate coding for machines. On the other hand, the trade-off between distortion and cognition (i.e., machine task performance) has not been publicly explored either.

\vspace{-2mm}
\subsection{Realism Image Compression}
\vspace{-1mm}
Traditional image coding goals often aim to achieve the lowest possible distortion. However, previous works \cite{blau2019rethinking} have demonstrated that "low distortion" is not synonymous with "high perceptual quality." There exists a trade-off among Rate, Distortion, and Realism (human perception). Consequently, GAN networks \cite{mentzer2020high} and Diffusion models \cite{yang2022lossy} have been employed to enhance realism. An attempt \cite{agustsson2023multi} was made to use a conditional factor to control the distortion-realism trade-off. EGIC \cite{korber2023egic} achieving similar effects using two decoders. Based on~\cite{agustsson2023multi}, Iwai \etal~\cite{iwai2024controlling} introduce a VBR mechanism to allow the control of rate-distortion-realism on a single codec.  Distinct from, yet akin to the Rate-Distortion-Realism framework, this paper considers the Rate-Distortion-Cognition issue, targeting the topic of image coding for machine (ICM).

\vspace{-2mm}
\section{Methodology}
\vspace{-2mm}
\subsection{Overview of the Proposed Compression Framework}
\vspace{-2mm}
\label{sec:frame}
\begin{figure}[t]
    \centering
    \includegraphics[width=1\linewidth]{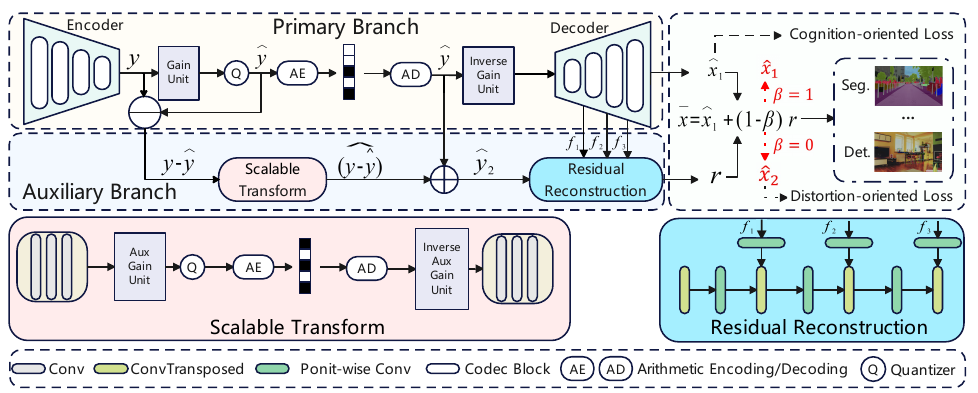}
    \vspace{-5mm}
    \caption{The framework of our method. The framework of our method. Our pipeline is divided into a primary branch and an auxiliary branch. The primary branch generates the cognition-oriented image $\hat{x}_1$, while the auxiliary branch generates the distortion-oriented residual $r$. We then use a $\hat{x}_1 + (1 - \beta)r$ interpolation strategy to achieve a balanced cognition-distortion trade-off. These two branches are trained separately in two stages: the primary branch uses cognition-oriented loss, and the auxiliary branch uses distortion-oriented loss.}
    \vspace{-4mm}
    \label{fig:framework}
\end{figure}

Our methodological approach is illustrated in Fig. \ref{fig:framework}. The training process is delineated into two stages, wherein the primary branch is dedicated to the generation of cognition-oriented compressed images in stage I, while the auxiliary branch focuses on the distortion-oriented compressed images in stage II. 

\begin{figure}[t]
    \centering
    \includegraphics[width=0.7\linewidth]{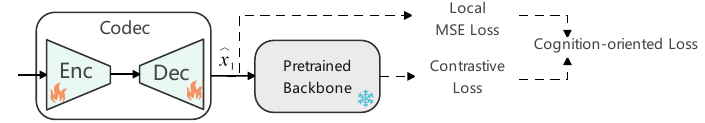}
    \vspace{-3mm}
    \caption{The training stage I. The parameters of the pretrained backbone remain frozen, with only the parameters of our codec being updated.}
    \vspace{-4mm}
    \label{fig:stageI}
\end{figure}

As shown in stage I of Fig. \ref{fig:stageI}, to enhance the performance of the compressed image $\boldsymbol{\hat{x}}_1$ on downstream machine vision tasks, we employ a cognition-oriented loss instead of the conventional distortion metric commonly used in image compression. This loss comprises two parts: 1). Inspired by studies~\cite{feng2022image,agustsson2019generative} indicating that semantic-level loss helps image cognition and downstream tasks in compression, the first component of the cognition-oriented loss designed herein encompasses a contrastive loss~\cite{he2020momentum} that facilitates the extraction of robust and discriminative representations.
2). Another part of the cognition-oriented loss is MSE of local regions in the reconstructed image where the pixel intensity exceeds [0,1]. This part is used as a penalty term to align the range of normal image values. 
The stage I process can be described as follows:
\vspace{-3mm}
\begin{equation}
\boldsymbol{x} \stackrel{encode}{\longrightarrow} \boldsymbol{y} \stackrel{quantization}{\longrightarrow} \boldsymbol{\hat{y}} \stackrel{decode}{\longrightarrow} \boldsymbol{\hat{x}}_1
\end{equation}
where $\boldsymbol{x}$ is the original image, $\boldsymbol{y}$ and $\boldsymbol{\hat{y}}$ are the latent code and quantized latent code extracted from $\boldsymbol{x}$. $\boldsymbol{\hat{x}}_1$ is the cognition-oriented compressed image. Furthermore, based on the existing structure, we further attain variable rate-cognition by regulating quantization degree using a channel-controllable latent gain unit~\cite{cui2020g}, this module is also trainable and compatible with the cognition-oriented loss.

In stage II, we implement the reconstruction of distortion-oriented compressed images to meet the requirements of human vision. Freezing the codec of the primary branch, we explore to generate distortion-oriented compressed images $\boldsymbol{\hat{x}}_2$ in this stage, which is defined as $\boldsymbol{\hat{x}}_2 = \boldsymbol{\hat{x}}_1 + \boldsymbol{r}$: the residual \(\boldsymbol{r}\) between the cognition-oriented and distortion-oriented compressed images is expected to extract from $\boldsymbol{\hat{y}}$ at the decoder side. However, since \(\boldsymbol{\hat{y}}\) is optimized for cognition in stage I, it didn't contain enough distortion-oriented supplemented information. Therefore, we construct an auxiliary branch to enrich the residual \(\boldsymbol{r}\) for the high-quality distortion-oriented compressed image $\boldsymbol{\hat{x}}_2$. As shown in Fig.~\ref{fig:framework}, we first encode the quantization error $(\boldsymbol{y - \hat{y}})$ at the encoder side for transmission. After a scalable transform with variable bitstream, the decoded error $(\boldsymbol{\widehat{y - \hat{y}}})$ is then added to the $\boldsymbol{\hat{y}}$ to get $\boldsymbol{\hat{y}}_2$, and a lightweight residual reconstruction network is used to refine residual $\boldsymbol{r}$. Interpolating $\boldsymbol{r}$ with $\boldsymbol{\hat{x}}_1$ using a weight $\beta$ gets $\boldsymbol{\bar{x}} = \boldsymbol{\hat{x}}_1 + (1 - \beta)\boldsymbol{r}$, which allows for different distortion-cognition performance. When $\beta$ equals 0, $\boldsymbol{\bar{x}}=\boldsymbol{\hat{x}_2}$ the lowest distortion is achieved; conversely, when $\beta$ equals 1, $\boldsymbol{\bar{x}}=\boldsymbol{\hat{x}_1}$ the highest cognition performance is attained.

\subsection{Primary Branch: Cognition-oriented Variable-bitrate Codec}
\label{sec;vbr}
A common use of learned image compression (LIC) model~\cite{liu2023learned,cheng2020learned,lu2021transformer,minnen2018joint}, can be formulated as the following equation:
\vspace{-3mm}
\begin{equation}
\begin{gathered}
\textbf{Encoding: }\boldsymbol{y}=g_{a}(\boldsymbol{x} ; \boldsymbol{\phi}_{g_a}) ;\quad \boldsymbol{\hat{y}}=Q(\boldsymbol{y})\\
\textbf{Decoding: } \boldsymbol{\hat{x}}=g_{s}(\boldsymbol{\hat{y}} ; \boldsymbol{\phi}_{g_s}) \\
\textbf{Hyper-prior: } \boldsymbol{z}=h_{a}\left(\boldsymbol{y} ; \boldsymbol{\phi}_{h_a}\right) ;\quad
\hat{\boldsymbol{z}}=Q(\boldsymbol{z}) ;\quad
\boldsymbol{\mu}_{y}, \boldsymbol{\sigma}_{y}=h_{s}\left(\boldsymbol{\hat{z}} ; \boldsymbol{\phi}_{h_s}\right)
\end{gathered}
\vspace{-2mm}
\label{equ:lic}
\end{equation}
where $\boldsymbol{\phi}_{g_a}, \boldsymbol{\phi}_{g_s}, \boldsymbol{\phi}_{h_a}$, and $\boldsymbol{\phi}_{h_s}$ are the parameters of the encoder network $g_a$, the decoder network $g_s$, the hyper-encoder network $h_a$, and the hyper-decoder network $h_s$, respectively. $Q$ represents the quantization which is approximated by a noise $\mathcal{U}\left(-\frac{1}{2}, \frac{1}{2}\right)$ during training, and is a rounding operator during inference. For inference, an image is first input to $g_a$ to extract a compact latent code $\boldsymbol{y}$, and $\boldsymbol{y}$ is quantized to obtain $\boldsymbol{\hat{y}}$ by quantization operator $Q$. On the decoder side, $\boldsymbol{\hat{y}}$ is input into $g_s$ for reconstruction to obtain the reconstructed image $\boldsymbol{\hat{x}}$. Additionally, the hyper path ($h_a$ and $h_s$) is used to help estimate the probability distribution of $\boldsymbol{y}$, which is assumed here to be a gaussian distribution with mean $\boldsymbol{\mu}_y$ and variance $\boldsymbol{\sigma}_y$.

\par
As discussed before, learned image compression typically requires learning a new model for different bitrate points. Therefore, in this paper, we employ a variable bitrate approach to achieve compression at different bitrates through a single model. Inspired by~\cite{cui2020g}, we scale $\boldsymbol{y}$ and $\boldsymbol{z}$ using $N$ gain unit $\boldsymbol{G}_{y,n}$ and $\boldsymbol{G}_{z,n}$, which is a matrix containing $C$ factors, where $C$ is the channel dimension of $\boldsymbol{y}$, and $n \in [1,N]$. By scaling different channels, we can adjust the precision of the data we need to transmit, further achieving transmission at $N$ different bitrates. Furthermore, we can interpolate the scaling factors for other bitrate points from these $N$ gain units, allowing us to switch between different bitrates using a single model. Therefore, Equation \ref{equ:lic} can be re-fomulated as:
\vspace{-2mm}
\begin{equation}
\begin{gathered}
\textbf{Encoding: }\boldsymbol{y}=g_{a}(\boldsymbol{x} ; \boldsymbol{\phi}_{g_a})  ;\quad \boldsymbol{\hat{y}}=Q(\boldsymbol{y} \cdot \boldsymbol{G}_{y,n}) / \boldsymbol{G}_{y,n} \\
\textbf{Decoding: }\boldsymbol{\hat{x}}=g_{s}(\boldsymbol{\hat{y}} ; \boldsymbol{\phi}_{g_s}) \\
\textbf{Hyper-prior: } \boldsymbol{z}=h_{a}\left(\boldsymbol{y}\cdot \boldsymbol{G}_{y,n} ; \boldsymbol{\phi}_{h_a}\right) ;\quad
\boldsymbol{\hat{z}}=Q(\boldsymbol{z}\cdot \boldsymbol{G}_{z,n}) / \boldsymbol{G}_{z,n} ;\\
\boldsymbol{\mu}_{y}, \boldsymbol{\sigma}_{y}=h_{s}\left(\boldsymbol{\hat{z}} ; \boldsymbol{\phi}_{h_s}\right)
\end{gathered}
\vspace{-2mm}
\label{equ:scale}
\end{equation}
For conventional distortion-oriented compression tasks, the loss can be represented as:
\vspace{-2mm}
\begin{equation}
\begin{aligned}
\mathcal{L}= & \mathcal{R}(\hat{\boldsymbol{y}})+\mathcal{R}(\hat{\boldsymbol{z}})+\lambda_n \cdot \mathcal{D}(\boldsymbol{x}, \hat{\boldsymbol{x}}) \\
= & \mathbb{E}\left[-\log _2\left(p_{\hat{\boldsymbol{y}} \mid \hat{\boldsymbol{z}}}(\hat{\boldsymbol{y}} \mid \hat{\boldsymbol{z}})\right)\right]+\mathbb{E}\left[-\log _2\left(p_{\hat{\boldsymbol{z}} \mid \boldsymbol{\psi}}(\hat{\boldsymbol{z}} \mid \psi)\right)\right] +\lambda_n \cdot \mathcal{D}(\boldsymbol{x}, \hat{\boldsymbol{x}})
\end{aligned}
\vspace{-2mm}
\end{equation}
where the parameter $\lambda_n$ regulates the balance between rate and distortion, with varying $\lambda_n$ values corresponding to different bit rates. $\boldsymbol{\psi}$ is a factorized density model used to encode quantized $\boldsymbol{\hat{z}}$. During training, we randomly select one of the $N$ bitrate points for each forward pass. The distortion term, denoted by $\mathcal{D}(\boldsymbol{x}, \boldsymbol{\hat{x}})$, is measured using the Mean Squared Error (MSE) loss, GAN loss or MS-SSIM loss. Additionally, $\mathcal{R}(\hat{\boldsymbol{y}})$ and $\mathcal{R}(\hat{\boldsymbol{z}})$ represent the bit rates for the latent representations $\hat{\boldsymbol{y}}$ and $\hat{\boldsymbol{z}}$, respectively.
To improve the performance of the reconstructed images for various downstream tasks, here we use cognition-oriented loss as a substitute for the original MSE loss~\cite{liu2023learned, cheng2020learned,lu2021transformer}. The first part of the cognition-oriented loss is a contrastive loss computed using a pretrained MoCo~\cite{he2020momentum}. It is defined as follows: 
\vspace{-3mm}
\begin{equation}
    \mathcal{C}=-\log \frac{\exp \left(q \cdot k_{+} / \tau\right)}{\sum_{i=0}^\mathtt{K} \exp \left(q \cdot k_i / \tau\right)}
    \vspace{-2mm}
\end{equation}
where $q$ denotes the query vector derived from the current image, $k_{+}$is a positive key vector generated from a different augmentation of the same image, and $k_i$ are the negative key vectors from different images. The denominator aggregates the exponentiated similarities of the query with both the positive and $\mathtt{K}$ negative keys, normalized by a temperature parameter $\tau$, which adjusts the sharpness of the distribution. This loss function encourages the model to produce embeddings where positive pairs are closer in the feature space than any negative pair, thus learning robust and discriminative feature representations. However, unlike previous work with contrastive loss, which typically aims to learn a better backbone~\cite{he2020momentum} or omni-features through contrastive learning~\cite{feng2022image}, our objective here is to update the codec's parameters to enable the generation of cognition-oriented compressed images. Therefore, we impose constraints on the output values of the codec, to align the range of normal image values and avoid outliers. Here, we compute a local Mean Squared Error (MSE) Loss for values lying outside the [0, 1] range. It can be formulated as:
\vspace{-3mm}
\begin{equation}
\mathcal{D}_{\text {local}}(\boldsymbol{x}, \hat{\boldsymbol{x}})=\frac{1}{M} \sum_{i=1}^N\left[\mathbb{I}\left(\hat{x}_i<0 \vee \hat{x}_i>1\right) \cdot\left(\hat{x}_i-x_i\right)^2\right]
\vspace{-2mm}
\end{equation}
Here, $\mathbb{I}$ is an indicator function that takes the value of 1 when the predicted value $\hat{x}_i$ falls outside the interval $[0,1]$, and 0 otherwise. $x_i$ and $\hat{x}_i$ represent the target and predicted values for the $i$ th pixel, respectively, while $M$ denotes the number of pixels. $\mathcal{D}_{\text {local}}$ is used as a penalty term to constrain the output range.

\begin{figure*}[t]
    \centering\includegraphics[width=1.\linewidth]{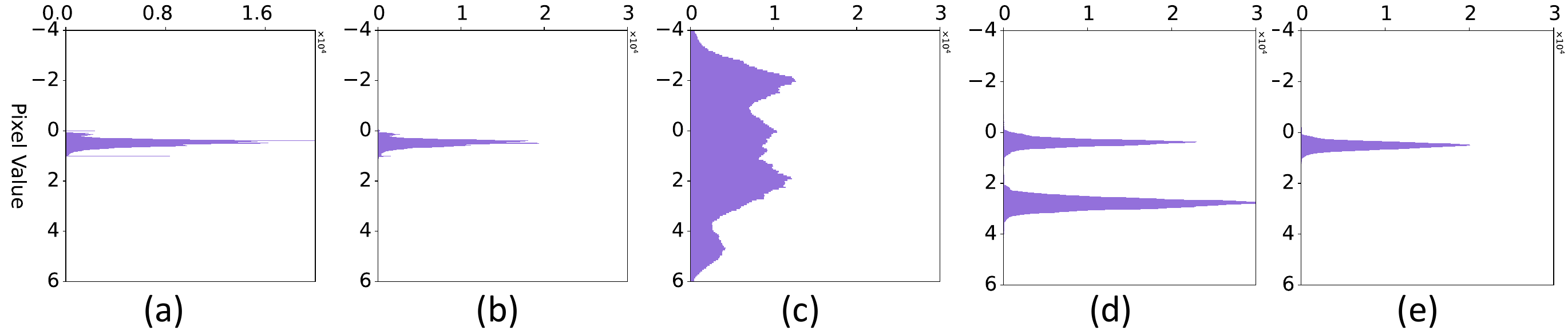}
\vspace{-6mm}
  \caption{ The histograms for pixel values of: (a) the original image. (b) distortion-oriented compressed image. (c) cognition-oriented compressed image without using local mse loss. (d) cognition-oriented compressed images processed with clipping to the range (0, 1) during training. (e) cognition-oriented compressed images with using local mse loss.}
\vspace{-3mm}
  \label{fig:distribution}
\end{figure*}

As illustrated in Fig. \ref{fig:distribution}, the image distributions of both the original image and the distortion-oriented compressed image primarily fall within the range of 0 to 1. However, if we directly optimize using the contrastive loss, the generated representations would reside within a significantly broader interval (approximately [-4, 6]), which does not align with the inherent characteristics of images. Moreover, directly clipping the generated representations to the [0, 1] range during training before passing them to subsequent networks can lead to the issue of vanishing gradients, resulting in a distribution of representations as shown in Fig. \ref{fig:distribution}(d). Clipping values outside the [0, 1] range during inference would lead to substantial information loss. Employing a local MSE loss as a penalty term can effectively mitigate this problem (see Fig. \ref{fig:distribution}(e)).

Therefore, the overall loss of stage I in the main branch is:
\vspace{-2mm}
\begin{equation}
\vspace{-1mm}
    \mathcal{L}_I= \mathcal{R}(\hat{\boldsymbol{y}})+\mathcal{R}(\hat{\boldsymbol{z}})+\lambda_n \cdot \mathcal{C}  + \lambda_{\text {local }} \cdot \mathcal{D}_{\text {local }}(\boldsymbol{x}, \hat{\boldsymbol{x}})
    \vspace{-2mm}
\end{equation}
where the parameter $\lambda_{\text {local}}$ serves to modulate the magnitude of influence exerted by the penalty term, denoted as $\mathcal{D}_{\text {local }}$. It is worth noting that the primary function of $\mathcal{D}_{\text {local}}$ loss within this context is to enforce constraints on the output range. Consequently, we empirically assign a relatively small fixed value of $10^{-5}$ to $\lambda_{\text {local}}$, acknowledging its role in merely constraining the output range without significantly impacting the primary loss components. 

\begin{table}[t]
\centering
\caption{Parameter numbers of different Codec modules.}
\vspace{-3mm}
\label{tab:param}
\resizebox{0.82\textwidth}{!}{\begin{tabular}{c|c|c|c}
\hline
\textbf{}               & \textbf{Base Codec} & \textbf{Scalable Transform} & \textbf{Residual Reconstruction} \\ \hline
\textbf{Parameters(/M)} & 20.39               & 0.19                        & 1.20                              \\ \hline
\end{tabular}}
\vspace{-4mm}
\end{table}
\vspace{-2mm}
\subsection{Auxiliary Branch: Distortion-oriented Residual Supplement}
\label{sec:distortion}

To further meet the requirements of human vision, we employ an auxiliary branch to supplement residual information with a scalable bitstream. As shown in Fig. \ref{fig:framework}, we construct a residual reconstruction network to generate the residual $\boldsymbol{r}$. By adding $\boldsymbol{r}$ to $\boldsymbol{\hat{x}}_1$, we reconstruct an distortion-oriented compressed image $\boldsymbol{\hat{x}}_2$. The residual reconstruction network comprises multiple layers of transposed convolutions and point-wise convolutional layers. The former is to increase the size of the representation, ensuring spatial dimension alignment with the original decoder's representation. The latter is employed to reduce and align the channel dimensions, simultaneously decreasing the model parameters and computational complexity. 
Additionally, the intermediate layer outputs $\boldsymbol{f_1, f_2}$, and $\boldsymbol{f_3}$ from the original decoder serve as prior information. They are fed into the residual reconstruction network through a point-wise convolution to help recovery. 

\begin{figure*}[t]
  \centering
 \begin{subfigure}{.31\linewidth}
    \centering\includegraphics[width=0.85\linewidth]{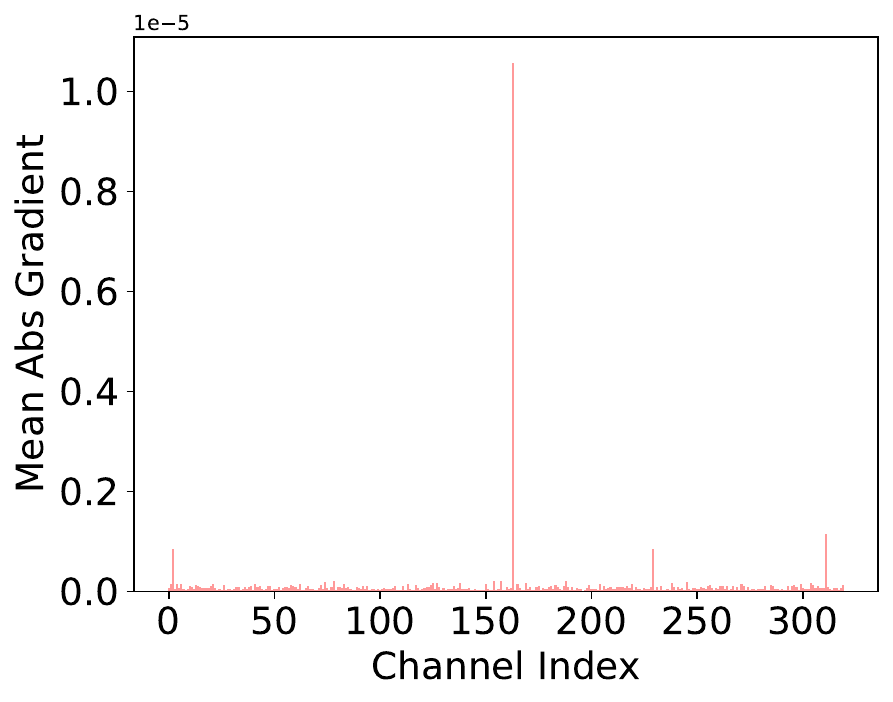}
  \end{subfigure}
  \begin{subfigure}{.31\linewidth}
    \centering\includegraphics[width=0.85\linewidth]{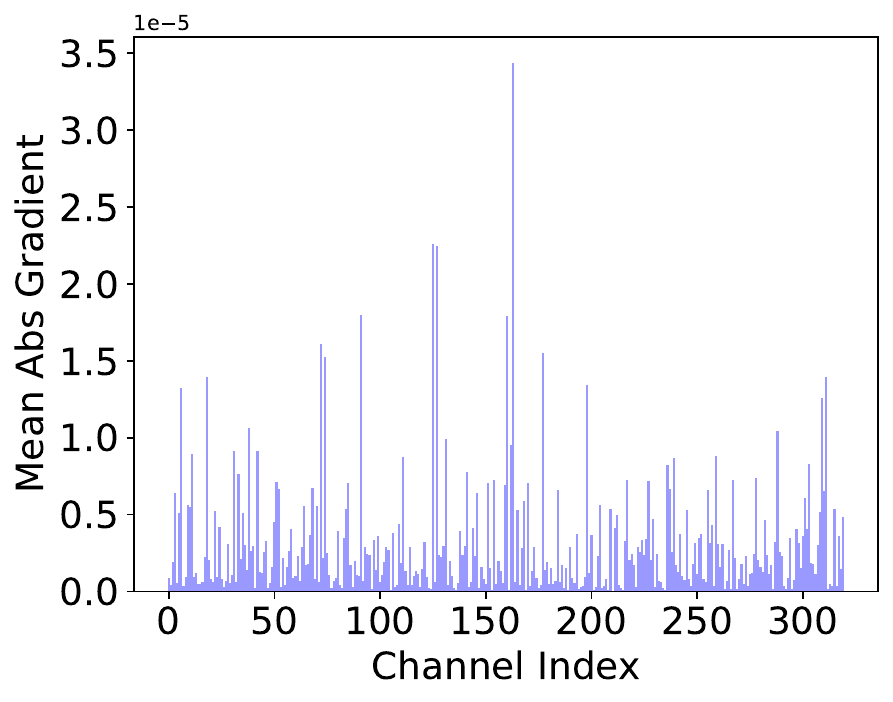}
  \end{subfigure}
  \begin{subfigure}{.33\linewidth}
    \centering\includegraphics[width=1.\linewidth]{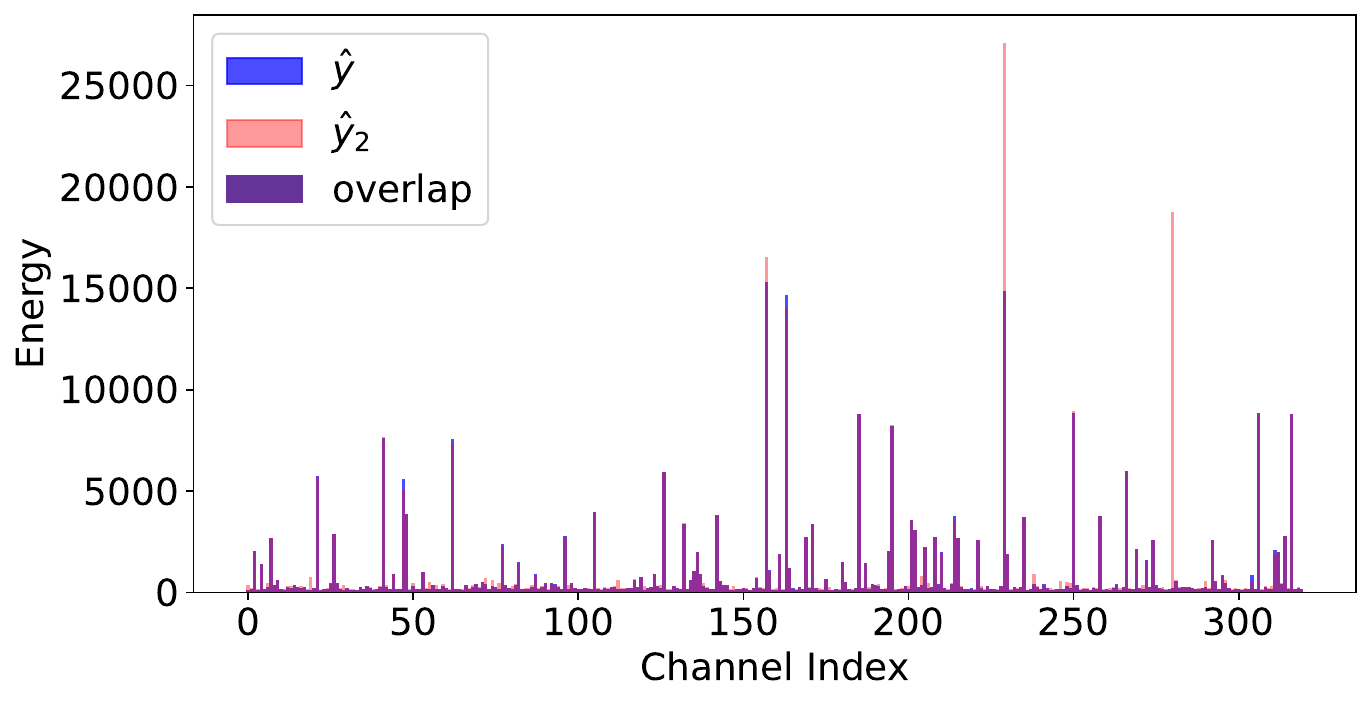}
  \end{subfigure}
    \vspace{-4mm}
  \caption{(a) The gradient for $\boldsymbol{\hat{y}}$ from cognition-orient image $\boldsymbol{\hat{x}}_1$ (b) The gradient for $\boldsymbol{\hat{y}}$ from distortion-orient image $\boldsymbol{\hat{x}}_2$ (c) The energy of cognition-orient $\boldsymbol{\hat{y}}$ and distortion-orient $\boldsymbol{\hat{y}}_2$ (Purple represents the overlapping part of the energy between \(\boldsymbol{\hat{y}}\) and \(\boldsymbol{\hat{y}_2}\)). }
  \vspace{-5mm}
  \label{fig:gradient}
\end{figure*}

We attempt to input $\boldsymbol{\hat{y}}$ to the residual reconstruction network directly to generate $\boldsymbol{r}$. However, directly using \({\boldsymbol{\hat{y}}}\) for reconstruction does not work, because $\boldsymbol{\hat{y}}$ of the primary branch is optimized for cognitive perception rather than distortion minimization. To prove that, we analyze the quantization operation: as shown in Equation \ref{equ:scale}, this process involves the multiplication of \(\boldsymbol{y}\) by \(\boldsymbol{G}_{y,n}\) followed by quantization. \(\boldsymbol{G}_{y,n}\) plays a crucial role in determining the precision of quantization across different channels and, i.e, the extent of quantization error. Given that \(\boldsymbol{G}_{y,n}\) is optimized for cognitive perception rather than distortion minimization in stage I, thus the reconstructed representations \(\boldsymbol{\hat{y}}\) cannot recover the detailed residual. The gradient visualization in Fig. \ref{fig:gradient}(a)(b) reveals the disparities in channel contributions between cognition-oriented and distortion-oriented images. By backpropagating the mean gradients $\frac{\partial \boldsymbol{\hat{x}}}{\partial \boldsymbol{\hat{y}}}$ from the generated image into $\boldsymbol{\hat{y}}$'s channels, we observe a marked difference in how $\boldsymbol{\hat{y}}$'s channels influence the generation of cognition-oriented versus distortion-oriented compressed images.

To address this issue, we propose to transmit this missed information through a scalable bitstream. Specifically, by using the similar VBR method in Sec. \ref{sec;vbr}, we first encode the quantization-induced error ($\boldsymbol{y-\hat{y}}$) into a bitstream via a transformation through three convolutional layers for transmission to the decoder side. An auxiliary gain unit is used to control the precision and bitrate of the bitstream. Then, we reconstruct bitstream through another three convolutional layers, and the output is added to $\boldsymbol{\hat{y}}$ to produce $\boldsymbol{\hat{y}}_2$, which is fed into the residual reconstruction network to generate $\boldsymbol{r}$. 
The scalable transform and the residual reconstruction network together form the auxiliary branch, which exclusively costs fewer parameters compared to the base codec (see Table \ref{tab:param}).

Moreover, we also analyze the differences in energy across various channels between $\boldsymbol{\hat{y}}_2$ and the original $\boldsymbol{\hat{y}}$ in Fig. \ref{fig:gradient}(c). It is observable that the energy levels across most channels are similar, with only a few channels significantly increased. This suggests that we only need to transmit very few bits to supplement such residuals to enhance decompression quality, which will be further proved in the experiments.
Therefore, the overall loss of stage II can be formulated as follows:
\vspace{-3mm}
\begin{equation}
\mathcal{L}_{II}=\mathcal{R}(\hat{\boldsymbol{s}})+\lambda_m \cdot \mathcal{D}(\boldsymbol{x}, \hat{\boldsymbol{x}}_2)
\end{equation}
where $\mathcal{R}(\hat{\boldsymbol{s}})$ represents the bitrate of the scalable bitstream, $\mathcal{D}$ is defined as MSE loss, $\lambda_m$ is used to control rate-distortion, $m \in [1,N]$, and $\boldsymbol{\hat{x}_2} = \boldsymbol{\hat{x}_1} + \boldsymbol{r}$. 

\vspace{-2mm}
\subsection{Rate-Distortion-Cognition Trade-off}
\vspace{-2mm}
Through the aforementioned procedures, we are able to achieve two distinct trade-offs: rate-distortion and rate-cognition. To achieve ultra-controllability, by employing interpolation, we can adjust the trade-off between distortion and cognition, thereby enabling the controllable rate-distortion-cognition. The equation is as follows:
\begin{equation}
\vspace{-3mm}
    \boldsymbol{\bar{x}} = \boldsymbol{\hat{x}_1} + (1-\beta) \times \boldsymbol{\hat{r}}
\end{equation}
where $\beta \in[0,1]$ is used to adjust the balance between cognition and distortion. As $\beta$ approaches 1, the distortion of $\boldsymbol{\bar{x}}$ increases, and the cognition performance of $\boldsymbol{\bar{x}}$ is enhanced. Conversely, as $\beta$ approaches 0 , the distortion decreases, and the cognition performance diminishs. When transferring to downstream tasks, $\beta$ is set to 1, the codec weights are kept frozen, and we only use cognition-oriented compressed images to finetune the network for the downstream tasks.

\vspace{-1mm}
\section{Experiments}
\vspace{-3mm}

\subsection{Experimental Setting}
\vspace{-2mm}
\subsubsection{Codec.}
We set TIC~\cite{lu2021transformer} as our base codec. During training stage I, at each step, we randomly select a number from the set [0.0625,0.125,0.25,0.5,1,2] as $\lambda_n$. Correspondingly, the latent code $\boldsymbol{y}$ is multiplied by the associated gain unit to implement variable bit rates. For convenience, in the subsequent sections, we use $\alpha \in[0,1]$ to represent the selected point of the bitrate. When $\alpha=0$, the bitrate is at its minimum, corresponding to the model with $\lambda_n= 0.0625$ . Conversely, when $\alpha=1$, the bitrate is at its maximum, corresponding to the model with $\lambda_n=2$. For training stage II, we set $\lambda_m \in [0.00018, 0.00036, 0.00072, 0.001, 0.0015, 0.002]$. To assess the rate-distortion performance of the reconstructed image, we employ the Peak Signal-to-Noise Ratio (PSNR) to evaluate the extent of distortion and use bits per pixel (bpp) to measure the compression ratio. Bpp is formulated as: as $\frac{b}{h \times w}$, where $h, w$ are the height and width of the source image, $b$ denotes the total bits cost of the coded feature bit-stream.
\vspace{-5mm}
\subsubsection{Image Classification.}
we train and validate our model on ImageNet-1k datasets. During the evaluation, before inputting the images into the model, we resize and crop them to 224x224. Meanwhile, ResNet50~\cite{he2016deep}, serving as the task network, is used to assess the accuracy of classification. 
\vspace{-5mm}
\subsubsection{Object Detection.}
For the object detection task, we train and validate on the MS COCO2017 train/val dataset. 
Following the setting in~\cite{wu2019detectron2}, we employ the faster-rcnn-FPN network~\cite{ren2015faster} for object detection. The image scale ranges from [640, 800] pixels during training and is set to 800 at inference by default.
\vspace{-5mm}
\subsubsection{Semantic Segmentation.}
 For this task, we employ DeepLab V3~\cite{chen2017deeplab}, a network designed for segmentation tasks. The training and evaluation are conducted on the Cityscapes dataset. The Cityscapes dataset is specifically designed for urban street scenes. It includes 5,000 images with fine annotations.
 Cross entropy serves as the loss function during the training process. To assess effectiveness, we use the mean intersection over union (mIoU) as the metric.
\vspace{-5mm}
\subsubsection{Baseline.}
For evaluating the effectiveness of our method on machine vision tasks, we have selected several different baselines for comparison with our approach. Initially, the traditional image standard VVC (VTM-12.0)~\cite{bross2021overview} was utilized for testing. For learned image compression, we employ TIC~\cite{lu2021transformer} to assess its performance across three tasks. Regarding the ICM method, we chose to compare with recent works, namely TransTIC~\cite{chen2023transtic} and T-TF~\cite{chamain2021end}. To evaluate the rate-distortion performance of our model, we conducted comparisons with methods such as VTM, TIC~\cite{lu2021transformer}, Factorized~\cite{balle2017end} and MBT-Mean~\cite{minnen2018joint}.

\vspace{-4mm}
\subsection{Quantitative Comparison}
\begin{figure}[t]
    \centering
    \includegraphics[width=1.\linewidth]{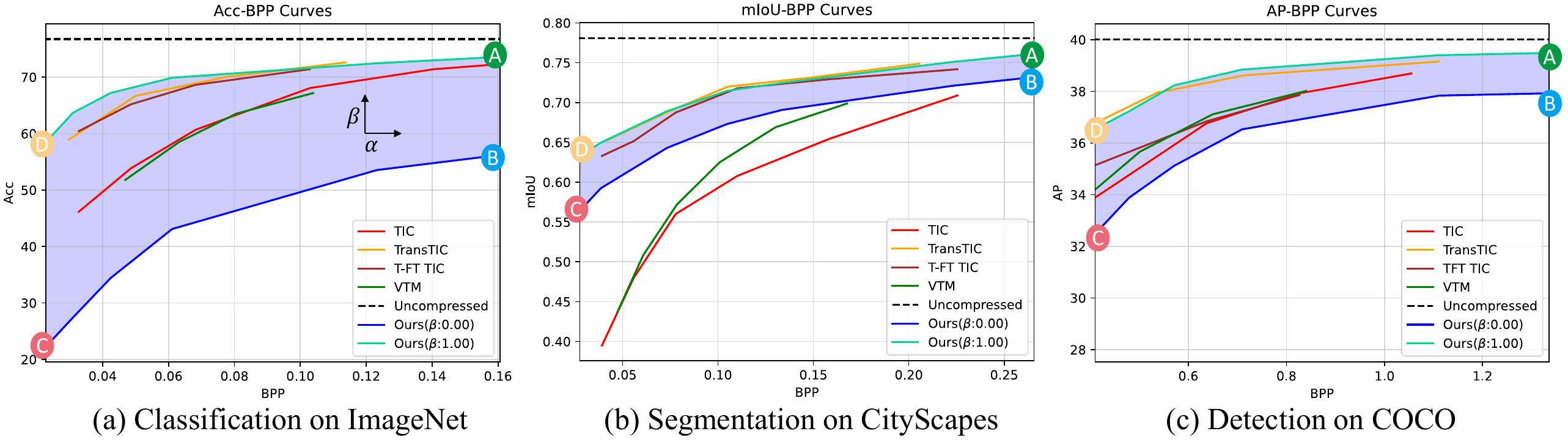}
    \vspace{-6mm}
    \caption{The rate-machine vision task performance curve ($\alpha$ changes from 0 to 1 from left to right, and $\beta$ changes from 0 to 1 from bottom to top).}
    \vspace{-4mm}
    \label{fig:result_mv}
\end{figure}

\begin{figure}[t]
    \centering
    \includegraphics[width=1.\linewidth]{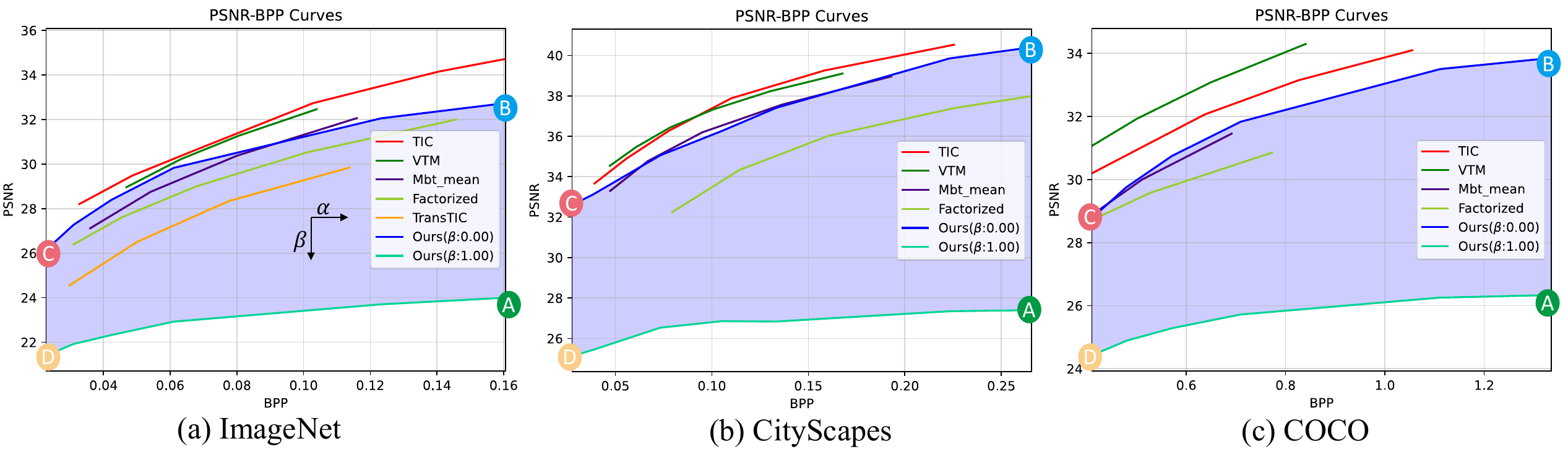}
    \vspace{-7mm}
    \caption{The rate-PSNR performance curve ($\alpha$ changes from 0 to 1 from left to right, and $\beta$ changes from 0 to 1 from top to bottom).}
    \vspace{-2mm}
    \label{fig:result_hv}
\end{figure}

\vspace{-2mm}
As illustrated in Fig. \ref{fig:result_mv} and \ref{fig:result_hv}, we compared our method against recent works. It is first noteworthy that our approach is controllable, by adjusting the values of $\alpha$ and $\beta$, we can achieve the ultra-controllability within the blue area depicted in the figures using a single codec. The bitrate increases with an increase in $\alpha$, while the performance for machine vision tasks and image reconstruction tasks respectively increases and decreases with an increase in $\beta$. As demonstrated by points ABCD in the Fig. \ref{fig:result_mv} and \ref{fig:result_hv}, at ($\beta=1$, $\alpha=1$) (\textcolor{ForestGreen}{Point A}), we achieve the highest bitrate and the optimal machine vision task performance at this bitrate ; at ($\beta=0$, $\alpha=1$) (\textcolor{NavyBlue}{Point B}), we obtain the highest bitrate and the lowest distortion at this bitrate; at ($\beta=0$, $\alpha=0$) (\textcolor{BrickRed}{Point C}), we achieve the lowest bitrate and the lowest distortion at this bitrate; at ($\beta=1$, $\alpha=0$) (\textcolor{Dandelion}{Point D}), we achieve the lowest bitrate and the best machine vision task performance at this bitrate.

As shown in Fig. \ref{fig:result_mv} and Table \ref{tab:result_mv}, we compared our method ($\beta=1$) with recent approaches in three tasks: classification, segmentation, and detection. It is evident that, compared with the un-tuned learned image compression method TIC~\cite{lu2021transformer} and the traditional image compression standard VVC (VTM-12.0)~\cite{bross2021overview}, our method significantly improves machine vision tasks performance, especially at lower bitrates. This demonstrates our model's superior cognition capabilities. Furthermore, compared with some ICM methods, TransTIC~\cite{chen2023transtic}, T-FT~\cite{chamain2021end} + TIC~\cite{lu2021transformer}, our approach not only achieves comparable or even superior performance but also provides controllability.

Fig. \ref{fig:result_hv} and Table \ref{tab:result_hv} illustrate the performance of our method ($\beta=0$) in the image reconstruction task. It is observed that compared with TIC, which is optimized for distortion, our method, being primarily optimized for cognition, leading to a reduction in PSNR. However, it still manages to achieve results comparable to the recent learned image compression methods, such as MBT-Mean~\cite{minnen2018joint}. Additionally, as demonstrated in Fig. \ref{fig:result_hv}(a), compared with the ICM method TransTIC, our method can achieve a higher upper bound for PSNR. 

\begin{table}[t]
\begin{minipage}[t]{0.49\linewidth} 
\centering
\caption{Comparison of machine vision task performance.}
\label{tab:result_mv}
\resizebox{0.82\textwidth}{!}{ 
\begin{tabular}{l|l|l|l}
\toprule
                  & \textbf{Cls.} & \textbf{Det.} & \textbf{Seg.} \\
\midrule
                  & \textbf{BD-Acc}         & \textbf{BD-AP}     & \textbf{BD-mIoU}      \\
\midrule
\textbf{TIC~\cite{lu2021transformer}}      &        0.6908                 & -0.4421            & -0.0208               \\
\midrule
\textbf{TransTIC~\cite{chen2023transtic}} &        8.9218                 & 1.8137             & 0.1128                \\
\midrule
\textbf{T-FT TIC~\cite{chamain2021end}} &        8.3459               & 0.7091             & 0.1049                \\
\midrule
\textbf{Ours ($\beta=1$)}     &             9.9876           & 1.8259             & 0.1133                 \\
\bottomrule
\end{tabular}
}
\end{minipage}
\hfill
\begin{minipage}[t]{0.49\linewidth} 
\centering
\caption{Comparison of image reconstruction performance.}
\resizebox{0.96\textwidth}{!}{ 
\begin{tabular}{l|l|l|l}
\toprule
                    & \textbf{Cls.} & \textbf{Det.} & \textbf{Seg.} \\
\midrule
                    & \textbf{BD-PSNR}        & \textbf{BD-PSNR}   & \textbf{BD-PSNR}      \\
\midrule
\textbf{TIC~\cite{lu2021transformer}}        &         0.2352                & -0.8876            & 0.0315                \\
\midrule
\textbf{MBT-Mean~\cite{minnen2018joint}}   &         -0.8492                & -2.1432            & -0.9654               \\
\midrule
\textbf{Factorized~\cite{balle2017end}} &         -1.5387                & -2.3248            & -3.4599               \\
\midrule
\textbf{Ours ($\beta=0$)}       &           -0.4102              & -1.8293            & -1.0240                \\
\bottomrule
\end{tabular}
}
\label{tab:result_hv}
\end{minipage}
\vspace{-3mm}
\end{table}
\vspace{-4mm}
\subsection{Ablation Study}
\begin{figure*}[t]
  \centering
 \begin{subfigure}{.45\linewidth}
    \centering\includegraphics[width=0.8\linewidth]{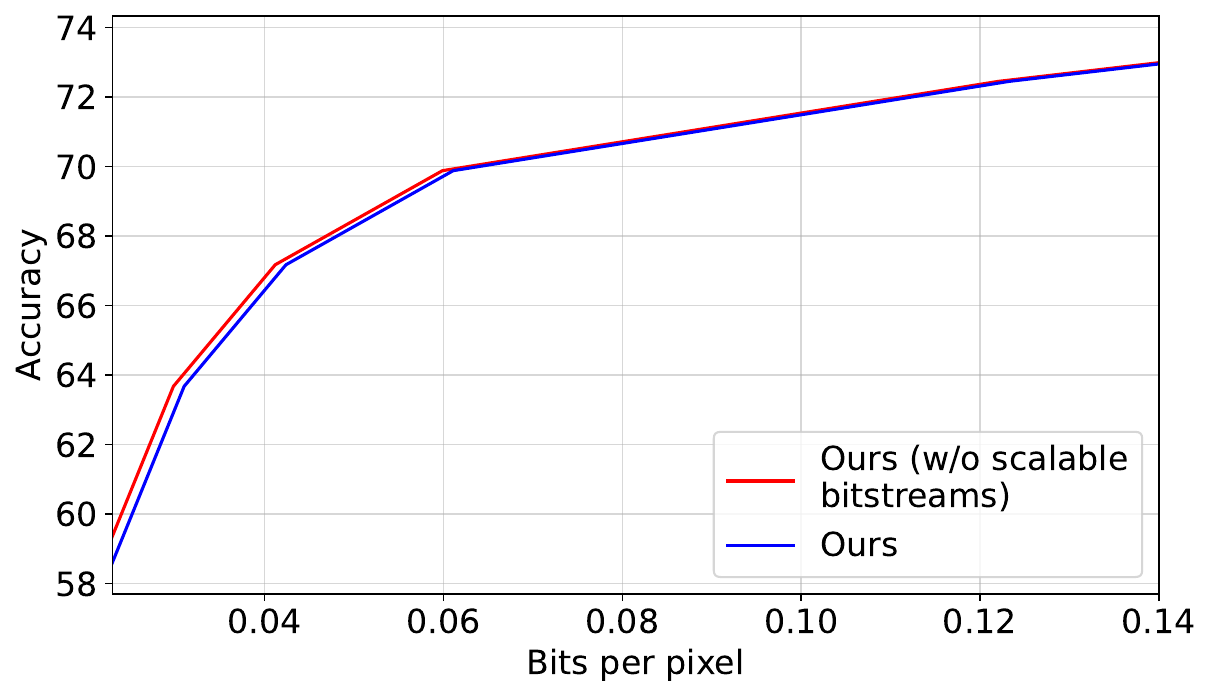}
    \caption{Rate-Acc.}
  \end{subfigure}
  \begin{subfigure}{.45\linewidth}
    \centering\includegraphics[width=0.8\linewidth]{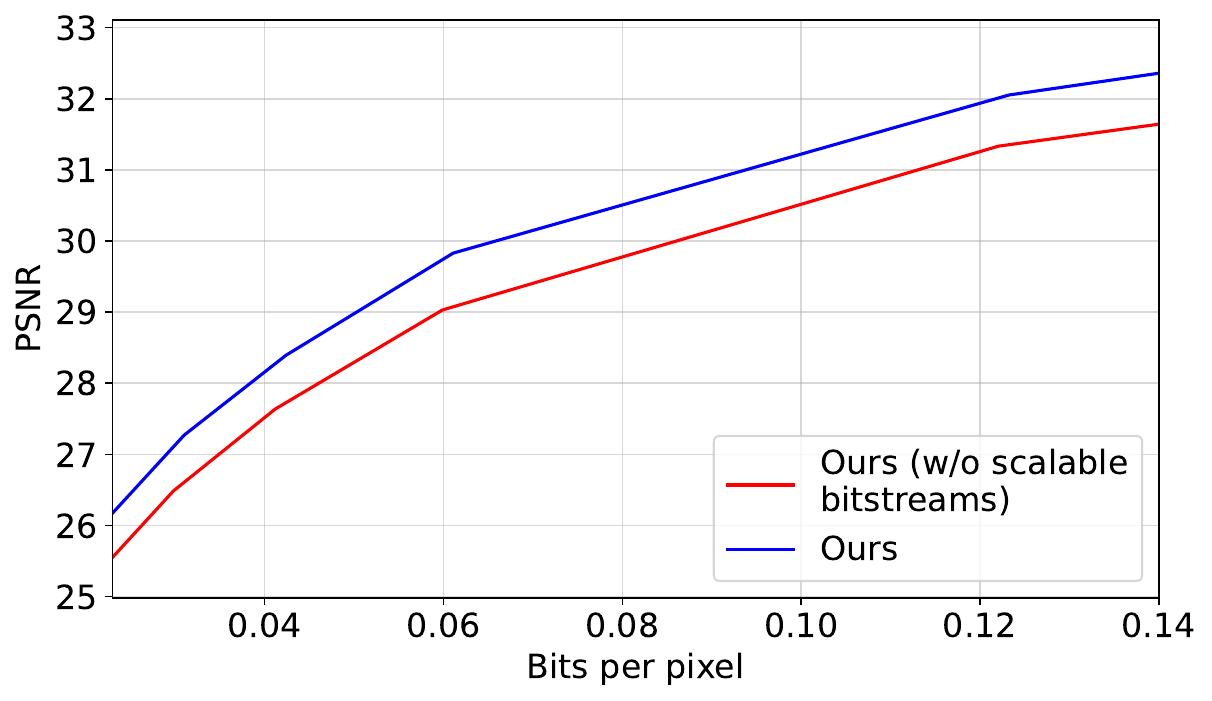}
    \caption{Rate-PSNR.}
  \end{subfigure}
    \vspace{-2mm}
  \caption{The ablation studies.}
  \vspace{-3mm}
  \label{fig:ablation}
\end{figure*}

\vspace{-2mm}
In our study, we conducted ablation studies based on the base codec TIC, as shown in Table \ref{tab:ablation}. For the measurements of BD-Acc~\cite{bdrate} and BD-PSNR, we set $\beta$=1 and $\beta$=0, respectively. We explored the application of the VBR method from~\cite{cui2020g} to TIC, which allows for varying bitrates with a single model. Directly transferring VBR methods to machine vision tasks does not yield satisfactory results. At the same time, for other distinct scenarios (different tasks/networks and cognition-distortion trade-off), codec retraining remains requisite. Using the proposed cognition-oriented loss can significantly improve the performance of machine vision tasks. However, for different bitrates and cognition-distortion trade-offs, many codecs are still required. In contrast, our method requires only one codec to be applicable across all scenarios. For our method, when we do not transmit an auxiliary scalable bitstream and directly use $\boldsymbol{\hat{y}}$ for the training of residual reconstruction, there is a significant improvement in BD-Acc, but the distortion-oriented compressed images exhibit heavy distortion. 
As analyzed in Section \ref{sec:distortion} and shown in Fig. \ref{fig:ablation} and Table \ref{tab:ablation}, we discovered that a minimal scalable bitstream can significantly reduce reconstruction distortion, with BD-PSNR improving by approximately 37.81\%, and the loss in BD-Acc being only around 2\%.
\begin{table}[t]
\centering
\caption{BD-PSNR and BD-Acc on ImageNet with TIC as anchor. Codec Number accounts for required codecs across $K$ bitrates, $M$ downstream tasks, and $T$ distortion-cognition scenarios.}
\vspace{-2mm}
\resizebox{0.96\textwidth}{!}{\begin{tabular}{c|c|c|c|c|c}
\hline
\textbf{Methods}       & \textbf{TIC~\cite{lu2021transformer}} & \textbf{\begin{tabular}[c]{@{}c@{}}TIC + \\ VBR~\cite{cui2020g}\end{tabular}}  & \textbf{\begin{tabular}[c]{@{}c@{}}TIC + \\ Cognition-oriented loss\end{tabular}} & \textbf{\begin{tabular}[c]{@{}c@{}}Ours \\ (w/o scalable bitstream)\end{tabular}} & \textbf{Ours} \\ \hline
\textbf{BD-PSNR}       & -            &    -0.1453                &  -1.5421           & -1.4640                                                                           & -0.9104       \\ \hline
\textbf{BD-Acc}        & -            &   -0.5739                &   11.1254          & 10.8124                                                                            & 10.6045        \\ \hline
\textbf{Codec Numbers} & $K\times M\times T$          & $M \times T$                 & $K \times T$                 & 1                                                                                 & 1             \\ \hline
\end{tabular}}
\vspace{-2mm}
\label{tab:ablation}
\end{table}

\vspace{-2mm}
\subsection{Visualization}
\begin{figure}[t]
    \centering
    \includegraphics[width=1\linewidth]{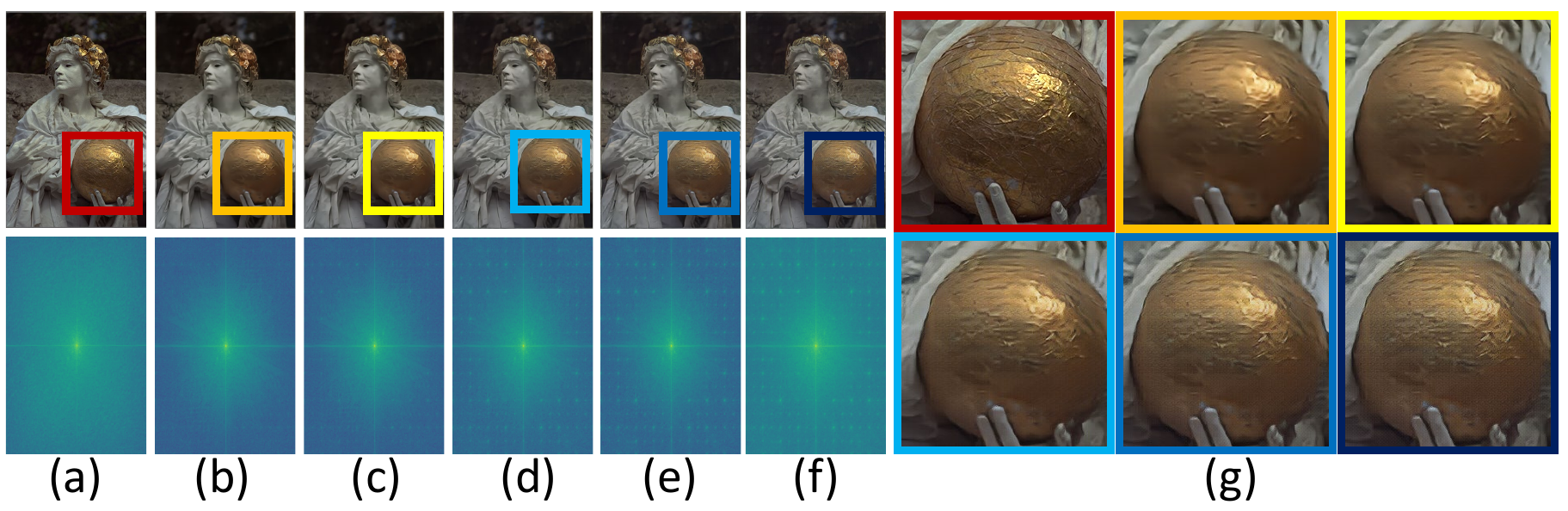}
    \vspace{-7mm}
    \caption{Visualization results and spectrum with different $\beta$ values. (a) represents the original image, while (b) to (f) represent the compressed images with $\beta$ values of 0, 0.25, 0.5, 0.75, and 1.0, respectively, (g) represents the magnification of the corresponding colored frame area.}
    \vspace{-3mm}
    \label{fig:visualization}
\end{figure}

\vspace{-2mm}
In Fig. \ref{fig:visualization}, we present a visualization of images generated with varying $\beta$ values, along with their spectrum. It is observable that distortion-oriented compressed images (i.e., $\beta=0$) lose a significant amount of high-frequency information compared to the original image. However, as the $\beta$ value increases, the high-frequency information in the images progressively increases. This indicates that cognition-oriented compressed images contain more high-frequency information, making them more conducive for analysis in downstream tasks. For instance, compared to distortion-oriented compressed images, cognition-oriented compressed images depict more detailed wrinkles on the ball held in the statue's hand. On the other hand, cognition-oriented compressed images, compared to both distortion-oriented compressed images and the original, possess a certain degree of color deviation. This suggests to some extent that color restoration is not crucial for some downstream tasks.
\vspace{-4mm}
\section{Conclusion}
\vspace{-2mm}
In this paper, we propose a Rate-Distortion-Cognition Controllable Versatile Neural Image Compression model. We introduce a cognition-oriented loss to enhance the machine vision support, and achieve variable-rate by latent channel regulation. To further meet human eye perception, we build an auxiliary branch with a very small and scalable bitrate, decoding distortion-oriented compressed images. Finally, with an interpolation strategy to merge two branches, we also achieve distortion-cognition controlling, thereby realizing controllable rate-distortion-cognition in a single codec. Our experiments demonstrate that our method can achieve excellent performance in classification, segmentation, and detection tasks while also achieving ultra-controllability.

\vspace{4mm}

\section*{Acknowledgments}

\noindent This work was supported in part by NSFC 62302246 and ZJNSFC under Grant LQ23F010008, and supported by High Performance Computing Center at Eastern Institute of Technology, Ningbo, and Ningbo Institute of Digital Twin.

\clearpage  

%
%
\bibliographystyle{splncs04}
\bibliography{main}
\end{document}